\documentclass[runningheads]{llncs}
\usepackage[T1]{fontenc}
\usepackage{amsmath}
\usepackage{xcolor}
\usepackage{booktabs}
\usepackage{multirow}
\usepackage{hyperref}
\usepackage{soul}

\definecolor{lb}{RGB}{230, 230, 255}
\definecolor{db}{RGB}{40, 40, 148}
\definecolor{ly}{RGB}{255, 255, 200} % Light yellow

\definecolor{darkgreen}{RGB}{0,130,0}

\soulregister\ref7
\soulregister\cite7
\soulregister\pageref7
\soulregister\citet7
\usepackage{graphicx,verbatim}
\usepackage{color}

\begin{document}
\title{RetFiner:
%An Efficient
A Vision-Language Refinement Scheme for Retinal Foundation Models}
\titlerunning{RetFiner}
\begin{comment}  %% Removed for anonymized MICCAI 2025 submission
\author{First Author\inst{1}\orcidID{0000-1111-2222-3333} \and
Second Author\inst{2,3}\orcidID{1111-2222-3333-4444} \and
Third Author\inst{3}\orcidID{2222--3333-4444-5555}}
%
\authorrunning{F. Author et al.}
\institute{Princeton University, Princeton NJ 08544, USA \and
Springer Heidelberg, Tiergartenstr. 17, 69121 Heidelberg, Germany
\email{lncs@springer.com}\\
\url{http://www.springer.com/gp/computer-science/lncs} \and
ABC Institute, Rupert-Karls-University Heidelberg, Heidelberg, Germany\\
\email{\{abc,lncs\}@uni-heidelberg.de}}

\end{comment}

\author{Ronald Fecso\inst{1,3} \and
José Morano\inst{1,2,3} \and
Ursula Schmidt-Erfurth\inst{4} \and
Hrvoje Bogunović\inst{1,2,3}}
% index{Fecso, Ronald}
% index{Morano, José}
% index{Schmidt-Erfurth, Ursula}
% index{Bogunović, Hrvoje}

\authorrunning{R. Fecso et al.}
\institute{Institute of Artificial Intelligence, Center for Medical Data Science, Medical University of Vienna, Vienna, Austria \\ \and Christian Doppler Lab for Artificial Intelligence in Retina, Center for Medical Data Science, Medical University of Vienna, Vienna, Austria \\ \and Comprehensive Center for AI in Medicine, Medical University of Vienna, Austria\\ \and 
OPTIMA Lab, Dept. of Ophthalmology, Medical University of Vienna, Austria\\
\email{\{ronald.fecso,jose.moranosanchez,hrvoje.bogunovic\}@meduniwien.ac.at}}

% \author{Anonymized Authors}  %% Added for anonymized MICCAI 2025 submission
% \authorrunning{Anonymized Author et al.}
% \institute{Anonymized Affiliations \\
%     \email{email@anonymized.com}}

\maketitle              % typeset the header of the contribution

\begin{abstract}
The rise of imaging techniques such as optical coherence tomography (OCT) and advances in deep learning (DL) have enabled clinicians and researchers to streamline retinal disease staging. A popular DL approach is self-supervised learning (SSL), where models learn from vast amounts of unlabeled data, avoiding costly annotation. SSL has allowed the development of foundation models (FMs), large models that can be used for a variety of downstream tasks.
However, existing FMs for OCT, trained solely on image data, lack a comprehensive and robust semantic understanding of images, as evidenced by their downstream performance (especially for complex tasks), and thus require supervised fine-tuning (which may be unfeasible) to better adapt to specific applications and populations. 
To address this, we propose RetFiner, an SSL vision-language refinement scheme that improves the representations of existing FMs and enables their efficient and direct adaptation to specific populations for improved downstream performance.
Our method uses a diverse set of training objectives which take advantage of the rich supervisory signal found in textual data. We tested RetFiner on the retinal FMs RETFound, UrFound, and VisionFM, showing significant improvements in linear probing performance on seven highly diverse OCT classification tasks, with an average increase of 5.8, 3.9, and 2.1 percentage points over their baselines, respectively. Our code and model weights are publicly available at \url{https://github.com/ronnief1/RetFiner}.


% Authors must provide keywords and are not allowed to remove this Keyword section.

\end{abstract}
\section{Introduction}

Ocular and systemic diseases affecting the eye represent an important health concern. Optical coherence tomography (OCT) has become the standard imaging technique to assess and diagnose several retinal diseases such as age-related macular degeneration (AMD)~\cite{Keenan2021}. %, which affects 200 million people~\cite{Keenan2021}.
With the emergence of OCT and other advanced imaging modalities, medical artificial intelligence (AI) offers great potential to accelerate the diagnostic process~\cite{zhou2023foundation}. However, traditional AI methods, mostly based on deep learning (DL), rely on large amounts of labeled data, which requires costly manual annotation. Recently, self-supervised learning (SSL) has gained popularity because it allows models to learn meaningful features from unlabeled data~\cite{ssl}. % which can then be applied to a variety of downstream tasks.
The combination of SSL techniques and large datasets and DL architectures has enabled the development of \textit{foundation models} (FMs)~\cite{bommasani2021opportunities}, generalizable models that can be efficiently adapted to several applications.

A common SSL approach is masked modeling (MM), which randomly masks part of the input and tasks the model with reconstructing the missing data, thus learning meaningful data representations.
The most common MM method for images is Masked Autoencoding (MAE)~\cite{he2022masked}, based on Vision Transformer (ViT)~\cite{dosovitskiy2020image}.

RETFound \cite{zhou2023foundation} applied MAE to develop separate FMs for retinal OCT and fundus images, demonstrating strong performance on diagnostic tasks.
% and generalizability.
UrFound \cite{urfound} trained an FM using joint MAE and masked language modeling (MLM).
Uni4Eye++~\cite{uni4eye} trained a model on multimodal data using a multi-step MAE approach.
Despite the promising results, some studies~\cite{maelimited2,balestriero2024how} have shown that MAE-based methods produce suboptimal representations for perceptual, highly semantic tasks.
% REFERENCES:
% - https://openreview.net/forum?id=XsDWw1Mn2p&referrer=%5Bthe%20profile%20of%20Yann%20LeCun%5D(%2Fprofile%3Fid%3D~Yann_LeCun1)
% - https://arxiv.org/pdf/2402.10093
In contrast, VisionFM~\cite{qiu2023visionfmmultimodalmultitaskvision} used a self-distillation approach to develop separate FMs for 8 ophthalmic modalities, resulting in improved performance over RETFound.
%In self-distillation, learning is achieved by feeding two different image views to two encoders, and mapping one to the other using a predictor.
Self-distillation learning consists of feeding two different image views to two encoders, and mapping one to the other using a predictor.

Regardless of the approach, existing FMs for OCT are still limited by the relatively small size of the pretraining dataset compared to FMs for computer vision~\cite{oquab2024dinov2learningrobustvisual,zhou2023foundation,qiu2023visionfmmultimodalmultitaskvision}, which leads to data bias, lower generalizability and, in some cases, the need for supervised tuning to specific populations or applications.

%\cite{eslami2021does,lin2023pmc,wang2022medclip,bioclip,zhang2022contrastive}
Another popular SSL approach is contrastive language-image pretraining (CLIP)~\cite{radford2021learning}, which consists of training a vision-language model (VLM) by aligning visual and textual representations from different encoders using an image-text contrastive (ITC) loss.
CLIP has gained popularity in the medical field due to the common availability of paired image and Electronic Health Record (EHR) data \cite{Silva_Rodr_guez_2025,zhao2024clipmedicalimagingcomprehensive}.
FLAIR \cite{Silva_Rodr_guez_2025} performed CLIP on 37 classification datasets of color fundus images by converting class labels into descriptions. However, this approach is not readily adaptable to unstructured data (e.g., EHRs) as it requires non-trivial decisions about how to create class labels and their descriptions.
%and \Js{2)} how to map those categories to captions following their knowledge-guidance framework.
Moreover, CLIP alone is suboptimal for
%developing medical vision-language models (VLMs)
medical data
because there exists high semantic overlap
%in medical data
(e.g., patients with the same diseases or biomarkers)~\cite{shui2025large}. This
%, in standard contrastive learning,
causes unpaired examples to be pushed apart in the embedding space regardless of their semantic similarity, resulting in false negatives \cite{fn}. Also, CLIP-based approaches usually struggle to distinguish subtle pathological patterns in medical images, as they rely only on global features~\cite{gloria}.
%Such limitations result in suboptimal downstream performance in the medical domain. 

% Some approaches have found that combining \J{instance-level?? What is?} instance-level and reconstruction-based tasks result in better representations  \cite{arici2021mlim,kwon2022masked}. 
To mitigate these issues, other works \cite{li2021align,yu2022coca,chen2023towards} have proposed to combine CLIP-like ITC losses with MM and image--text matching (ITM) objectives.
%, with improved downstream performance.
ALBEF~\cite{li2021align} trains a 3-encoder network (image, text, and multimodal) using ITC, ITM and MLM losses and a momentum model.
%However, their method requires an extra momentum model, resulting in longer and more expensive training. 
%COSA~\cite{chen2024cosa} 
In the medical domain, PTUnifier \cite{chen2023towards} used a similar setting but with a single model by unifying text and image inputs via prompts.
%to achieve SOTA performance on a range of cross-modal and multi-modal tasks . 
CoCa \cite{yu2022coca} achieved SOTA zero-shot classification performance of natural images by jointly training a VLM on contrastive and captioning tasks.
%Despite strong downstream performance, the previous methods do not fully utilize the rich signal available in image-text datasets yet still require long training.
Despite strong downstream performance, the use of these advanced approaches to develop or improve retinal FMs has not yet been explored.

\paragraph{Contribution.}
In this work, we present RetFiner (Fig.~\ref{Fig:OCTVLM}), an efficient vision-language \underline{re}finement scheme for \underline{re}tinal found\underline{a}tion mo\underline{d}els. Our approach consists of training a VLM composed of 
%two encoders (
a vision encoder based on an arbitrary retinal FM and a separate
%transformer-based
language model using a set of diverse training objectives focused on exploiting the EHRs as supervisory signals for improving visual representations.
% Specifically, the text encoder network includes a cross-attention mechanism to the visual features so that it can serve as a single-modality encoder, cross-modality encoder, and decoder.
To validate our approach, we refined the retinal FMs RETFound~\cite{zhou2023foundation}, UrFound~\cite{urfound}, and VisionFM~\cite{qiu2023visionfmmultimodalmultitaskvision} with our scheme using an in-house dataset of 100k pairs of OCTs and associated EHRs.
Running RetFiner on this dataset for an FM requires less than 10 epochs.
% takes less than an hour on a single NVIDIA A100 GPU (80GB).
Linear probing of the refined vision FMs on six public and one in-house OCT classification datasets demonstrates the effectiveness of the proposed approach for both improving the semantic understanding of the models and adapting them to targeted populations.

Our RetFiner models set a new benchmark for OCT image analysis, with potential applications where high-level semantics are required, such as visual question aswering.
%To facilitate reproducibility and research progress, we have made the code and model weights publicly available.
To facilitate
%reproducibility and
research progress, we have published the code and model weights.

%=======================================================================
%=======================================================================
%=======================================================================
\section{Methods}

\begin{figure}[btp]
    \centering
    \includegraphics[scale=.5]{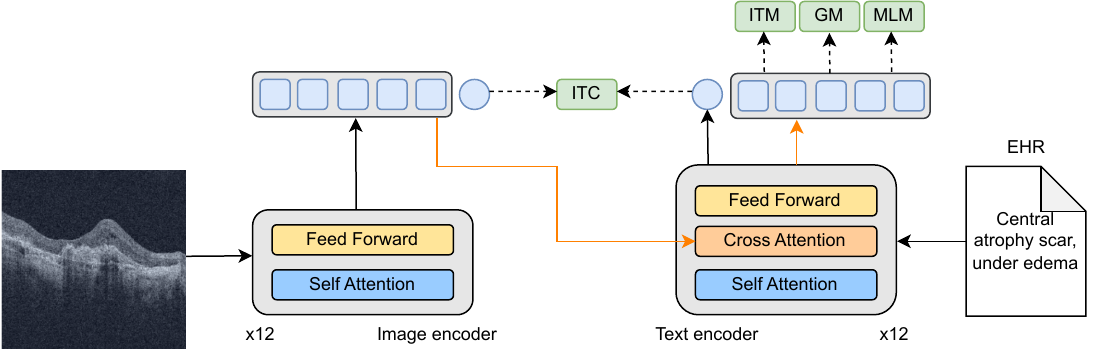}
    \caption{RetFiner method. Squares represent patch features and circles represent global features (CLS tokens). Cross-attention layers are activated only during the forward passes for ITM, MLM, and GM. An example of an OCT image and report is shown.}%
    \label{Fig:OCTVLM}
\end{figure}

RetFiner, as seen in Fig.~\ref{Fig:OCTVLM}, employs a simple architecture comprised of a ViT~\cite{dosovitskiy2020image} vision encoder and a Transformer text encoder~\cite{NIPS2017_3f5ee243}. This architecture allows single-modality and cross-modality embedding. Cross-attention (CA) layers are added between the self-attention and feed-forward layers of the text encoder.
%The cross-attention layers
CA layers are utilized for cross-modality encoding and generation, but deactivated for uni-modal encoding. The text encoder therefore acts as a uni-modal or cross-modal encoder as well as a decoder.

The model is trained to optimize four losses, as shown in Fig.~\ref{Fig:OCTVLM}: image-text contrastive (ITC), image-text matching (ITM), masked language modeling (MLM), and generative modeling (GM) (captioning). Such a combination effectively enhances the model’s cross-modality alignment, understanding, and generation capabilities, ultimately improving visual representations.
The total loss is a direct sum of all the losses equally weighted:
\begin{equation}
\mathcal{L}=\mathcal{L}_{\mathrm{ITC}}+\mathcal{L}_{\mathrm{ITM}}+\mathcal{L}_{\mathrm{MLM}}+\mathcal{L}_{\mathrm{GM}}.
\end{equation}

To efficiently use the model for downstream classification tasks, we propose a simple feature pooling strategy that integrates both global and local features. In particular, we concatenate the CLS token and the average pool of the patch tokens. Then, these features are fed into a trainable linear layer.

%=======================================================================
%=======================================================================

%=======================================================================
%=======================================================================
\subsection{Training Objectives}

\textit{~~~~ITC.}
We employ an ITC loss to align the vision and text encoders in the embedding space in order to learn better multi-modal representations. We use the InfoNCE loss from \cite{oord2019representationlearningcontrastivepredictive} to bring pairs of images and texts together in the embedding space while pushing apart negative pairs.

\textit{ITM.}
We use an ITM loss along with ITC to further align the image and text encoders. To challenge the model during training, we use hard negative mining, inspired by \cite{li2021align}. In this scheme, for a given image, the model must predict whether another report from the same batch is part of the same pair. The candidate report is sampled from other reports in the batch, where reports with higher cosine similarity in the embedding space to the given image have a higher chance of being sampled. This loss is also calculated with images and reports swapping places. Such a loss forces the model to learn to differentiate highly similar samples and in turn learn more discriminative features.

\textit{MLM.}
Despite the success of
%self-supervised
instance-level tasks such as ITC and ITM, they may be limited by the semantic overlap in medical images~\cite{zhao2024clipmedicalimagingcomprehensive}. To address this, we include two reconstruction tasks: MLM~\cite{devlin2018bert} and GM. These non-contrastive tasks act as regularizers for
% the instance-level contrastive tasks
ITC and ITM, while improving the model's semantic understanding and generation.
%capabilities.
MLM aims to predict randomly masked input tokens from reports in a bidirectional manner using Cross-Entropy (CE) loss.

\textit{GM.} To supplement the MLM task and enhance the model's reconstruction and generation abilities, we add a generative text modeling task. Given the context of an OCT and previous report tokens, the model is trained to auto-regressively predict the next masked report token using CE as loss function.

%=======================================================================
%=======================================================================
\subsection{Development Data}

For model development, we used an in-house dataset of 100k paired OCT images and EHRs, from which we extracted the text describing the OCT scan. The reports cover a range of retinal conditions such as cataracts, choroidal neovascularization (CNV), age-related macular degeneration (AMD), retinal vein occlusion (RVO), and glaucoma.
In addition, we used an extra 160k images with no EHRs to train a MAE baseline for the ablation studies.
All scans were collected between 2007 and 2021 at the *** Clinic at the University of ***. Images were taken with Cirrus and Spectralis devices.
In line with previous work~\cite{zhou2023foundation}, only the central B-scans of the 3D OCT volumes were used in this study.

\subsection{Implementation Details}
The pipeline is implemented in Python 3.10 using PyTorch \cite{NEURIPS2019_9015} and trained on a NVIDIA A100 GPU (80GB). The AdamW optimizer was used with a learning rate of $10^{-4}$ and a batch size of 128. Early stopping was triggered when the validation loss did not decrease for three epochs.
The vision encoder is based on the ViT architecture \cite{dosovitskiy2020image} and text encoder is a pre-trained BERT \cite{devlin2018bert}.
During the forward pass for the ITC loss, cross-attention was turned off in the text encoder. The CLS tokens from each encoder were projected down to a dimension of 512 with a linear layer, then L2-normalized before being passed into the ITC loss. For the remaining losses, cross-attention was activated between the self-attention and feed forward layers of each block. The patch tokens from the vision encoder were passed into the text encoder as the hidden states to perform cross-attention. For the MLM loss, 15\% of the report tokens are randomly masked. For the GM loss, 60\% of the report tokens are masked with causal attention masks.

%=======================================================================
%=======================================================================
\section{Experiments and Results}

\begin{table}[b]
\centering
\caption{Average linear probing performance over all downstream datasets. We compare the best metric out of all the models (bolded) with the best metric out of the base models (underlined) to measure if there was a statistically significant difference using the Wilcoxon signed-rank test
(**: \textit{p} < 0.01, ***: \textit{p} < 0.001). Values in parentheses represent change in performance compared to their baseline counterpart. }
\setlength{\tabcolsep}{3pt} % Adjust column spacing
%\resizebox{\textwidth}{!}{
{\fontsize{8}{10}\selectfont
\begin{tabular}{l l l l}
\hline
\textbf{{Model}} & \textbf{BAcc (\%)} & \textbf{AUROC (\%)} & \textbf{AP (\%)} \\
\hline
CLIP & 73.6$\pm$12.3 & 90.6$\pm$8.6 & 84.5$\pm$9.9 \\
DINOv2 & 74.5$\pm$12.5 & 90.4$\pm$9.7 & 84.7$\pm$11.7 \\
\midrule
RETFound & 78.1$\pm$11.6 & \underline{92.9}$\pm$7.7 & 88.6$\pm$9.5 \\
UrFound & 78.6$\pm$11.1 & 92.7$\pm$8.5 & 88.5$\pm$9.5 \\
VisionFM & \underline{81.4}$\pm$13.4 & 92.8$\pm$9.1 & \underline{89.3}$\pm$10.7 \\
\midrule
RetFiner-R & \textbf{83.8***}$\pm$12.0 \textcolor{darkgreen}{\text{(+5.8)}} & 94.6$\pm$6.7 \textcolor{darkgreen}{\text{(+1.6)}} & 91.2$\pm$9.5 \textcolor{darkgreen}{\text{(+2.6)}} \\
RetFiner-U & 82.5$\pm$12.3 \textcolor{darkgreen}{\text{(+3.9)}} & \textbf{94.7**}$\pm$7.7 \textcolor{darkgreen}{\text{(+1.9)}} & \textbf{92.3***}$\pm$8.4 \textcolor{darkgreen}{\text{(+3.9)}} \\
RetFiner-V & 83.5$\pm$12.2 \textcolor{darkgreen}{\text{(+2.1)}} & 94.4$\pm$7.5 \textcolor{darkgreen}{\text{(+1.6)}} & 91.2$\pm$10.2 \textcolor{darkgreen}{\text{(+1.9)}} \\
\hline
\end{tabular}}
\label{tab:model_resultsnoduke}
\end{table}

\begin{table}[p]
\caption{Linear probing performance on downstream datasets. We compare SOTA FMs for OCT with our RetFiner-refined versions of them. Performance differences are shown in parentheses. For each metric, we compare the best overall result (bold) with the best result of the base models (underlined) using Student's $t$-test (*: \textit{p} < 0.05, **: \textit{p} < 0.01, ***: \textit{p} < 0.001). 
The number of test cases ($n$) and classes ($C$) are also listed.}
\label{tab:sota1}
%\resizebox{\textwidth}{!}{%
\centering
{\fontsize{8}{10}\selectfont
\begin{tabular}{l@{\hskip 5pt}l@{\hskip 5pt}l@{\hskip 5pt}l@{\hskip 5pt}l}
\toprule
\textbf{Dataset} & \textbf{Model} & \textbf{BAcc (\%)} & \textbf{AUROC (\%)} & \textbf{AP (\%)} \\ 
\midrule
   \multirow{6}{*}{\shortstack{\textbf{In-house}\\$n=640$\\~\\~\\staging \&\\diagnosis\\$C=9$}}
        
        & RETFound & 75.3$\pm$2.3 \ & \underline{98.7}$\pm$0.2 \ & 92.0$\pm$0.7 \\ 
        %& Uni4Eye++ & 48.30 $\pm$ 2.73 \ & 91.01 $\pm$ 1.25 \ & 69.88 $\pm$ 2.92 \\ 
        & UrFound & 76.3$\pm$0.5 & 98.6$\pm$0.1 & \underline{92.5}$\pm$0.4 \\
        & VisionFM & \underline{80.1}$\pm$1.4 \ & 98.3$\pm$0.1 \ & 91.5$\pm$0.5 \\ 
   \cmidrule{2-5}
        & RetFiner-R  & \textbf{84.3***}$\pm$1.1 \textcolor{darkgreen}{\text{(+9.1)}} \ & \textbf{99.1***}$\pm$0.1 \textcolor{darkgreen}{\text{(+0.5)}} \ & \textbf{95.2***}$\pm$0.1 \textcolor{darkgreen}{\text{(+3.3)}}  \\ 
        %& RetFiner-U  & 74.07 $\pm$ 0.29 \textcolor{darkgreen}{\text{(+25.78)}} \ & 97.74 $\pm$ 0.05 \textcolor{darkgreen}{\text{(+6.73)}} \ & 88.85 $\pm$ 0.30 \textcolor{darkgreen}{\text{(+18.97)}}  \\ 
        & RetFiner-U  & 81.9$\pm$1.2 \textcolor{darkgreen}{\text{(+5.6)}} & 99.0$\pm$0.1 \textcolor{darkgreen}{\text{(+0.4)}} & 94.2$\pm$0.3 \textcolor{darkgreen}{\text{(+1.8)}}  \\ 
        & RetFiner-V  & 82.2$\pm$1.0 \textcolor{darkgreen}{\text{(+2.1)}} \ & 98.4$\pm$0.1 \textcolor{darkgreen}{\text{(+0.1)}} \ & 93.0$\pm$0.5 \textcolor{darkgreen}{\text{(+1.5)}}  \\ 

   \midrule
   \multirow{6}{*}{\shortstack{\textbf{GAMMA} \\ $n=20$\\~\\~\\staging \&\\diagnosis\\$C=3$}} 
        
        & RETFound & 54.7$\pm$9.3 \ & 80.2$\pm$0.5 \ & 69.3$\pm$2.5 \\ 
        %& Uni4Eye++ & 51.39 $\pm$ 3.93 \ & 79.10 $\pm$ 0.51 \ & \underline{69.56} $\pm$ 1.49 \\ 
        & UrFound & \underline{57.0}$\pm$3.9 & \underline{82.3}$\pm$0.7 & \underline{72.0}$\pm$3.3 \\ 
        & VisionFM & 53.6$\pm$6.0 \ & 77.0$\pm$4.2 \ & 68.1$\pm$5.1 \\ 
   \cmidrule{2-5}
        & RetFiner-R  & 58.9$\pm$7.5 \textcolor{darkgreen}{\text{(+4.2)}} \ & 84.6$\pm$1.9 \textcolor{darkgreen}{\text{(+4.5)}} \ & 71.8$\pm$3.9 \textcolor{darkgreen}{\text{(+2.6)}}  \\ 
        %& RetFiner-U  & 58.06 $\pm$ 6.09 \textcolor{darkgreen}{\text{(+6.67)}} \ & \textbf{84.71***} $\pm$ 1.72 \textcolor{darkgreen}{\text{(+5.62)}} \ & \textbf{78.07***} $\pm$ 3.25 \textcolor{darkgreen}{\text{(+8.52)}}  \\ 
        & RetFiner-U  & \textbf{60.0}$\pm$6.1 \textcolor{darkgreen}{\text{(+3.1)}} & \textbf{88.2***} $\pm$0.9 \textcolor{darkgreen}{\text{(+5.9)}} & \textbf{80.1***}$\pm$1.7 \textcolor{darkgreen}{\text{(+8.1)}}  \\
        & RetFiner-V  & 59.4$\pm$2.9 \textcolor{darkgreen}{\text{(+5.8)}} \ & 82.2$\pm$2.2 \textcolor{darkgreen}{\text{(+5.1)}} \ & 70.4$\pm$3.6 \textcolor{darkgreen}{\text{(+2.3)}}  \\ 
       %  & RetFiner-B & 56.94 $\pm$ 5.56 \ & 82.51 $\pm$ 0.69 \ & 71.45 $\pm$ 1.48 \\ 
   \midrule
   \multirow{6}{*}{\shortstack{\textbf{Harvard}\\\textbf{Glaucoma}\\$n=400$\\~\\~\\diagnosis\\$C=2$}} 
        
        & RETFound & \underline{74.6}$\pm$1.8 \ & \underline{82.0}$\pm$0.8 \ & \underline{80.8}$\pm$0.5 \\ 
        %& Uni4Eye++ & 62.17 $\pm$ 1.34 \ & 68.25 $\pm$ 0.50 \ & 67.94 $\pm$ 1.27 \\ 
        & UrFound & 71.2$\pm$1.4 & 77.3$\pm$0.6 & 76.6$\pm$0.7 \\ 
        & VisionFM & 74.5$\pm$0.7 \ & 81.2$\pm$0.6 \ & 80.7$\pm$0.6 \\ 
   \cmidrule{2-5}
        & RetFiner-R  & \textbf{77.7**}$\pm$0.6 \textcolor{darkgreen}{\text{(+3.1)}} \ & \textbf{83.8**}$\pm$0.4 \textcolor{darkgreen}{\text{(+1.7)}} \ & 83.1$\pm$0.4 \textcolor{darkgreen}{\text{(+2.3)}}  \\ 
        %& RetFiner-U  & 64.85 $\pm$ 2.16 \textcolor{darkgreen}{\text{(+2.68)}} \ & 71.34 $\pm$ 1.21 \textcolor{darkgreen}{\text{(+3.09)}} \ & 71.97 $\pm$ 1.10 \textcolor{darkgreen}{\text{(+4.03)}}  \\ 
        & RetFiner-U  & 70.1$\pm$0.6 \textcolor{red}{\text{(--1.1)}} & 78.6$\pm$0.4 \textcolor{darkgreen}{\text{(+1.2)}} & 79.0$\pm$0.3 \textcolor{darkgreen}{\text{(+2.4)}}  \\ 
        & RetFiner-V  & 74.5$\pm$1.3 (0.0) \ & 83.4$\pm$0.8 \textcolor{darkgreen}{\text{(+2.2)}} \ & \textbf{83.3***}$\pm$0.9 \textcolor{darkgreen}{\text{(+2.6)}}  \\ 
        % & RetFiner-B & 72.52 $\pm$ 1.26 \ & 79.23 $\pm$ 0.92 \ & 78.91 $\pm$ 0.78 \\ 
   \midrule
   \multirow{6}{*}{\shortstack{\textbf{NEHUT} \\ $n=135$\\~\\~\\diagnosis\\$C=3$}} 
        
        & RETFound & 84.7$\pm$1.2 \ & 95.2$\pm$0.5 \ & \underline{91.9}$\pm$0.8 \\ 
        %& Uni4Eye++ & 76.82 $\pm$ 0.69 \ & 88.69 $\pm$ 0.83 \ & 78.97 $\pm$ 1.66 \\ 
        & UrFound & 84.8$\pm$1.3 & 95.2$\pm$0.4 & 90.9$\pm$1.1 \\ 
        & VisionFM & \underline{88.2}$\pm$1.4 \ & \underline{95.3}$\pm$0.7 \ & 91.8$\pm$1.5 \\ 
   \cmidrule{2-5}
        & RetFiner-R  & \textbf{89.5*}$\pm$0.6 \textcolor{darkgreen}{\text{(+4.8)}} \ & 97.6$\pm$0.2 \textcolor{darkgreen}{\text{(+2.4)}} \ & 95.6$\pm$0.3 \textcolor{darkgreen}{\text{(+3.7)}}  \\ 
        %& RetFiner-U  & \textbf{89.77*} $\pm$ 1.12 \textcolor{darkgreen}{\text{(+12.95)}} \ & 96.95 $\pm$ 0.08 \textcolor{darkgreen}{\text{(+8.25)}} \ & 93.97 $\pm$ 0.18 \textcolor{darkgreen}{\text{(+15.00)}}  \\ 
        & RetFiner-U  & 89.5$\pm$0.6 \textcolor{darkgreen}{\text{(+4.7)}} & \textbf{97.7***}$\pm$0.2 \textcolor{darkgreen}{\text{(+2.5)}} & \textbf{96.0***}$\pm$0.5 \textcolor{darkgreen}{\text{(+5.1)}}  \\ 
        & RetFiner-V  & 88.0$\pm$0.8 \textcolor{red}{\text{(--0.2)}} \ & 97.5$\pm$0.1 \textcolor{darkgreen}{\text{(+2.3)}} \ & 95.5$\pm$0.3 \textcolor{darkgreen}{\text{(+3.7)}}  \\ 
        % & RetFiner-B & 87.87 $\pm$ 0.84 \ & 97.00 $\pm$ 0.17 \ & 94.80 $\pm$ 0.40 \\ 

    \midrule
   \multirow{6}{*}{\shortstack{\textbf{Noor Eye}\\\textbf{Hospital}\\$n=30$\\~\\~\\diagnosis\\$C=3$}} 
   
        & RETFound & 88.0$\pm$3.8 \ & 97.6$\pm$0.2 \ & 96.2$\pm$0.4 \\ 
        & UrFound & 88.0$\pm$5.1 & 97.6$\pm$0.4 & 96.0$\pm$0.6 \\ 
        & VisionFM & \underline{94.0}$\pm$2.8 \ & \underline{98.8}$\pm$0.1 \ & \underline{98.0}$\pm$0.1 \\ 
   \cmidrule{2-5}
        & RetFiner-R  & \textbf{95.3}$\pm$1.8 \textcolor{darkgreen}{\text{(+7.3)}} \ & 97.7$\pm$0.3 \textcolor{darkgreen}{\text{(+0.1)}} \ & 96.5$\pm$0.4 \textcolor{darkgreen}{\text{(+0.3)}}  \\ 
        %& RetFiner-U  & 88.00 $\pm$ 2.98 \textcolor{darkgreen}{\text{(+6.67)}} \ & 97.73 $\pm$ 0.30 \textcolor{darkgreen}{\text{(+3.77)}} \ & 96.38 $\pm$ 0.60 \textcolor{darkgreen}{\text{(+4.09)}}  \\ 
        & RetFiner-U  & 92.0$\pm$1.8 \textcolor{darkgreen}{\text{(+4.0)}} & 99.8$\pm$0.2 \textcolor{darkgreen}{\text{(+2.2)}} & 99.6$\pm$0.4 \textcolor{darkgreen}{\text{(+3.6)}}  \\ 
        & RetFiner-V  & 93.3$\pm$0.0 \textcolor{red}{\text{(--0.7)}} \ & \textbf{99.8***}$\pm$0.1 \textcolor{darkgreen}{\text{(+1.0)}} \ & \textbf{99.6***}$\pm$0.1 \textcolor{darkgreen}{\text{(+1.6)}}  \\ 
       %  & RetFiner-B & 93.33 $\pm$ 0.00 \ & 98.33 $\pm$ 0.33 \ & 97.18 $\pm$ 0.42 \\ 
   \midrule
   \multirow{6}{*}{\shortstack{\textbf{OCTDL}\\$n=332$\\~\\~\\diagnosis\\$C=7$}} 
   
        & RETFound & 80.3$\pm$1.2 \ & 97.9$\pm$0.3 \ & 93.6$\pm$0.6 \\ 
        %& Uni4Eye++ & 57.53 $\pm$ 3.34 \ & 91.94 $\pm$ 0.82 \ & 79.32 $\pm$ 1.35 \\ 
        & UrFound & 84.4$\pm$0.9 & 99.0$\pm$0.0 & 95.4$\pm$0.2 \\ 
        & VisionFM & \underline{87.6}$\pm$3.4 \ & \underline{99.2}$\pm$0.1 \ & \underline{96.5}$\pm$0.6 \\ 
   \cmidrule{2-5}
        & RetFiner-R  & 87.9$\pm$2.0 \textcolor{darkgreen}{\text{(+7.6)}} \ & 99.5$\pm$0.0 \textcolor{darkgreen}{\text{(+1.6)}} \ & 97.1$\pm$0.2 \textcolor{darkgreen}{\text{(+3.5)}}  \\ 
        %& RetFiner-U  & 80.14 $\pm$ 1.19 \textcolor{darkgreen}{\text{(+22.61)}} \ & 98.00 $\pm$ 0.18 \textcolor{darkgreen}{\text{(+6.06)}} \ & 93.44 $\pm$ 0.24 \textcolor{darkgreen}{\text{(+14.12)}}  \\ 
        & RetFiner-U  & 90.6$\pm$1.3 \textcolor{darkgreen}{\text{(+6.2)}} & \textbf{99.5***}$\pm$0.0 \textcolor{darkgreen}{\text{(+0.6)}} & \textbf{98.4***}$\pm$0.1 \textcolor{darkgreen}{\text{(+3.0)}}  \\ 
        & RetFiner-V  & \textbf{90.9*}$\pm$1.5 \textcolor{darkgreen}{\text{(+3.3)}} \ & 99.4$\pm$0.2 \textcolor{darkgreen}{\text{(+0.2)}} \ & 97.7$\pm$0.3 \textcolor{darkgreen}{\text{(+1.2)}}  \\ 
       % & RetFiner-B & 87.15 $\pm$ 1.01 \ & 99.28 $\pm$ 0.02 \ & 97.14 $\pm$ 0.13 \\ 
   \midrule
   \multirow{6}{*}{\shortstack{\textbf{OCTID}\\$n=174$\\~\\~\\diagnosis\\$C=5$}} 
   
        & RETFound & 88.8$\pm$2.0 \ & 99.0$\pm$0.1 \ & 96.7$\pm$0.2 \\ 
        %& Uni4Eye++ & 60.72 $\pm$ 3.47 \ & 92.55 $\pm$ 0.86 \ & 76.69 $\pm$ 2.46 \\ 
        & UrFound & 88.7$\pm$2.5 & 98.9$\pm$0.2 & 96.0$\pm$0.5 \\ 
        & VisionFM & \underline{91.8}$\pm$0.9 \ & \underline{99.7}$\pm$0.1 \ & \underline{98.6}$\pm$0.2 \\ 
   \cmidrule{2-5}
        & RetFiner-R  & 93.2$\pm$1.8 \textcolor{darkgreen}{\text{(+4.4)}} \ & 99.7$\pm$0.1 \textcolor{darkgreen}{\text{(+0.6)}} \ & 98.9$\pm$0.3 \textcolor{darkgreen}{\text{(+2.2)}}  \\ 
        %& RetFiner-U  & 85.77 $\pm$ 0.51 \textcolor{darkgreen}{\text{(+25.05)}} \ & 98.95 $\pm$ 0.09 \textcolor{darkgreen}{\text{(+6.39)}} \ & 95.58 $\pm$ 0.34 \textcolor{darkgreen}{\text{(+18.89)}}  \\ 
        & RetFiner-U  & 93.4$\pm$1.3 \textcolor{darkgreen}{\text{(+4.7)}} & 99.8$\pm$0.1 \textcolor{darkgreen}{\text{(+0.8)}} & 99.0$\pm$0.3 \textcolor{darkgreen}{\text{(+2.9)}}  \\ 
        & RetFiner-V  & \textbf{96.3***}$\pm$1.1 \textcolor{darkgreen}{\text{(+4.5)}} \ & \textbf{99.8***}$\pm$0.0 \textcolor{darkgreen}{\text{(+0.1)}} \ & \textbf{99.2***}$\pm$0.1 \textcolor{darkgreen}{\text{(+0.6)}}  \\ 
\bottomrule
\end{tabular}}
\end{table}

\paragraph{Experimental Setup.}
%\subsection{Comparison to the State of the Art}
To validate our approach, we applied RetFiner on our OCT--text dataset (100k) to three SOTA FMs: \underline{R}ETFound~\cite{zhou2023foundation}, \underline{U}rFound \cite{urfound}, and \underline{V}isionFM \cite{qiu2023visionfmmultimodalmultitaskvision}, resulting in \mbox{RetFiner-[\underline{R},\underline{U},\underline{V}]} refined models.
In addition, for the ablation studies, to ensure that the efficacy of our approach is not derived from our imaging data alone and to discard data bias, we used the full dataset (260k OCTs) to pretrain a ViT-Base model using MAE \cite{he2022masked}, as in \cite{zhou2023foundation}, and then applied RetFiner for comparison.
The performance of the refined models was then compared with that of the out-of-the-box models and two SOTA general-purpose vision models: CLIP~\cite{radford2021learning} and DINOv2~\cite{oquab2024dinov2learningrobustvisual}.
The performance was evaluated via linear probing with our concatenation pooling strategy on seven retinal disease classification datasets, namely OCTDL \cite{Kulyabin2024-gd}, OCTID \cite{octid}, GAMMA \cite{WU2023102938}, Harvard Glaucoma \cite{glauc}, NEHUT \cite{SOTOUDEHPAIMA2022105368}, Noor Eye Hospital \cite{8166817}, and an in-house dataset [anon. ref.]. These datasets cover a range of demographics, devices, and diseases, including AMD, glaucoma, diabetic retinopathy and diabetic macular edema. 
Each experiment was run five times with different seeds.
Models were evaluated using balanced accuracy (BAcc), area under receiver operating characteristic curve (AUROC), average prevision (AP), and, for ablation, also F1.

\paragraph{State-of-the-art Comparison.}
%We first compared our method with SOTA retinal and general-purpose FMs on the seven downstream datasets.
Table~\ref{tab:model_resultsnoduke} shows the average performance of our method and the SOTA retinal and general-purpose FMs across the different tasks.
Per-dataset performances are shown in Table~\ref{tab:sota1}. As shown in the tables, the top three performances across all metrics come from models refined using our proposed approach, with RetFiner-R and RetFiner-U performing the best in terms of BAcc and AUROC and AP, respectively.

\paragraph{Improvement Analysis.} % Used paragraph to save space
Tables~\ref{tab:model_resultsnoduke} and~\ref{tab:sota1} show a significant improvement in downstream classification performance for RetFiner models compared to their off-the-shelf counterparts. These results demonstrate the effectiveness of our approach for improving existing FMs.
This is more remarkable considering our method requires less than ten epochs to refine a model. 
Also importantly, our method significantly improves all retinal FMs on our complex in-house dataset (with 9 classes), which represents pathologies of high clinical relevance in general and in our clinic in particular. This demonstrates our methods's ability to mitigate the data bias found in retinal FMs while also leveraging their powerful representations, allowing us to create a model suitable for our applications using in-house data, with no need for manual annotation or data processing.

\begin{table}[tbp]
\centering
\caption{Linear probing performance of combinations of losses on our in-house dataset.}
\setlength{\tabcolsep}{3pt} % Adjust column spacing
%\resizebox{\textwidth}{!}{
{\fontsize{8}{10}\selectfont
\begin{tabular}{l l l l l}
\hline
\textbf{Losses} & \textbf{BAcc (\%)} & \textbf{AUROC (\%)} & \textbf{AP (\%)} & \textbf{F1-score (\%)} \\
\hline
ITC (\raisebox{0.29ex}{$\sim$}
CLIP \cite{radford2021learning,Silva_Rodr_guez_2025}) & 78.4$\pm$0.7 & 98.4$\pm$0.1 & 91.6$\pm$0.4 & 85.6$\pm$1.0 \\
MLM (\raisebox{0.29ex}{$\sim$}
UrFound \cite{urfound}) & 77.0$\pm$1.5 & 98.4$\pm$0.1 & 92.4$\pm$0.6 & 84.8$\pm$0.7 \\
\cmidrule{1-5}
ITC+GM & 80.0$\pm$0.8 & \underline{98.6}$\pm$0.1 & \underline{93.4}$\pm$0.1 & 86.5$\pm$0.3 \\
ITC+GM+MLM & \underline{80.1}$\pm$0.4 & 98.6$\pm$0.1 & 92.7$\pm$0.1 & \underline{86.7}$\pm$0.6 \\
ITC+GM+MLM+ITM & \textbf{82.0***}$\pm$0.7 & \textbf{98.7**}$\pm$0.1 & \textbf{93.5}$\pm$0.1 & \textbf{88.2***}$\pm$0.3 \\
\hline
\end{tabular}}
\label{tab:loss_ablate}
\end{table}

\begin{table}[tbp]
\centering
\caption{Linear probing performance of each pooling strategy on our in-house dataset.}
\setlength{\tabcolsep}{6pt} % Adjust column spacing
%\resizebox{\textwidth}{!}{
{\fontsize{8}{10}\selectfont
%{\fontsize{8}{10}\selectfont
\begin{tabular}{l l l l l}
\hline
\textbf{Model} & \textbf{BAcc (\%)} & \textbf{AUROC (\%)} & \textbf{AP (\%)} & \textbf{F1-score (\%)} \\
\hline
CLS token & 79.2$\pm$1.3 & 98.5$\pm$0.1 & 92.6$\pm$0.6 & 86.4$\pm$1.3 \\
Patch features & 79.3$\pm$0.2 & \underline{98.8}$\pm$0.0 & \underline{93.3}$\pm$0.1 & 87.3$\pm$0.3 \\
All tokens & \underline{79.5}$\pm$0.9 & \textbf{98.8}$\pm$0.0 & 93.3$\pm$0.1 & \underline{87.5}$\pm$0.4 \\
Concatenation & \textbf{82.0***}$\pm$0.7 & 98.7$\pm$0.1 & \textbf{93.5**}$\pm$0.1 & \textbf{88.2**}$\pm$0.3 \\
\hline
\end{tabular}}
\label{tab:token_ablate}
\end{table}

\paragraph{Baseline Comparison and Ablation Study.}
We compared our method to existing VLM approaches and tested the effect of the different losses on downstream performance to demonstrate their positive effect.
In both cases, we used our MAE-pretrained model as the base model and passed it through our scheme using the loss combinations listed in Table \ref{tab:loss_ablate}.
CLIP and UrFound baselines are analogous to training using only ITC and MLM losses, respectively.
The resulting models were evaluated by linear probing on our in-house dataset.
%, which has nine classes, the most of the downstream datasets.
As shown in Table \ref{tab:loss_ablate}, all losses result in significantly higher downstream performance.
We also tested the effect of our token pooling strategy for linear probing. During refinement, the ITC loss uses only the CLS token, while the other losses use only the patch tokens. We asserted that using both 
%the CLS and patch tokens
led to a better exploitation of the models' features.
%To test this, we again used the MAE trained on OCTs and passed it through our scheme. We evaluated by linear probing on our in-house dataset.
We compared this strategy with other common pooling strategies: CLS token, average pooling of patch tokens, and average pooling of all tokens (patch+CLS). Table \ref{tab:token_ablate} shows that the concatenation technique results in a statistically significant increase in balanced accuracy, AP, and F1-score.

\begin{figure}[tb]
    \centering
    \includegraphics[width=0.95\linewidth]{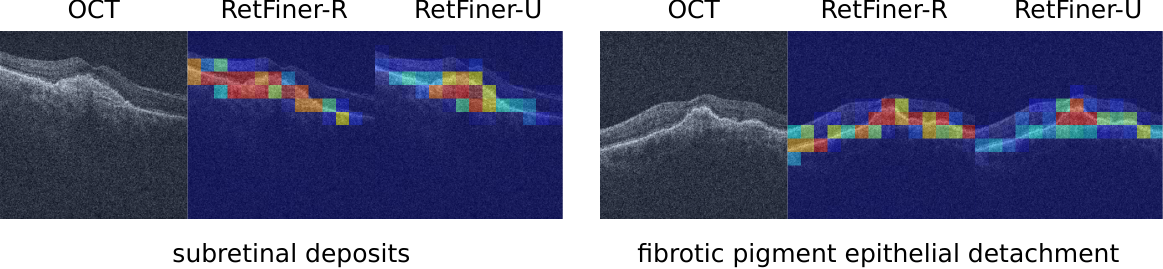}
    \caption{Examples of RetFiner-R and RetFiner-U attention maps for disease cases.}
    \label{Fig:att_maps}
\end{figure}

\paragraph{Explainability.}
Fig.~\ref{Fig:att_maps} shows examples of the attention rollout maps for our refined models RetFiner-R and RetFiner-U, based on RETFound and UrFound, respectively. The attention maps are generated by calculating the cross-attentions in the text encoder between the image features and the text features.
As shown in the images, RetFiner's maps highlight the retinal layers and activate more strongly around the lesion biomarkers indicated in the text, regardless of the base FM.
This demonstrates the effective semantic understanding of OCT images by our RetFiner models and their explanatory capabilities.

\section{Conclusion}

We introduced RetFiner, a vision-language refinement scheme that enhances the semantic understanding of retinal FMs through SSL on paired OCT images and EHR text. By combining multiple training objectives,
%(image-text contrastive, image-text matching, masked language modeling, and generative modeling)
RetFiner leverages textual data to refine visual representations without manual annotation or processing effort. Evaluated on seven diverse classification tasks, RetFiner significantly improved the linear probing performance of SOTA retinal FMs (RETFound, UrFound, VisionFM), achieving average gains of up to 5.8 percentage points in balanced accuracy. Notably, RetFiner demonstrated strong adaptability to our complex in-house dataset,
% (9-class staging/diagnosis),
highlighting its utility for population-specific adaptation. Ablation studies confirmed the effectiveness of each training objective and our feature pooling strategy. Furthermore, attention visualizations revealed that RetFiner models focus on clinically relevant biomarkers, enhancing explainability. With efficient training (under 10 epochs) and compatibility with existing FMs, RetFiner offers a practical solution to adapt models to local data distributions while improving overall semantic understanding and performance. Notably, our scheme is not specific to ophthalmic data and could be easily applied to other medical domains where paired image-text data is available. 

\begin{credits}
\subsubsection{\ackname} This research was funded in part by the Austrian Science Fund (FWF) Grant-DOI:10.55776/FG9, Christian Doppler Research Association, Austrian Federal Ministry of Economy, Energy and Tourism, and the National Foundation for Research, Technology and Development. For open access purposes, the author has applied a CC BY public copyright license to any author-accepted manuscript version.

\subsubsection{\discintname}
The authors have no competing interests to declare that are relevant to the content of this article.
\end{credits}
\begin{comment}  %% removed for anonymized MICCAI 2025 submission.
    
    % The following acknowledgement and disclaimer sections should be removed for the double-blind review process.  
    % If and when your paper is accepted, reinsert the acknowledgement and the disclaimer clause in your final camera-ready version.

\begin{credits}
\subsubsection{\ackname} A bold run-in heading in small font size at the end of the paper is
used for general acknowledgments, for example: This study was funded
by X (grant number Y).

\subsubsection{\discintname}
It is now necessary to declare any competing interests or to specifically
state that the authors have no competing interests. Please place the
statement with a bold run-in heading in small font size beneath the
(optional) acknowledgments\footnote{If EquinOCS, our proceedings submission
system, is used, then the disclaimer can be provided directly in the system.},
for example: The authors have no competing interests to declare that are
relevant to the content of this article. Or: Author A has received research
grants from Company W. Author B has received a speaker honorarium from
Company X and owns stock in Company Y. Author C is a member of committee Z.
\end{credits}

\end{comment}


%\clearpage

\bibliographystyle{splncs04}
\bibliography{bibliography}
%





% LNCS Homepage, \url{http://www.springer.com/lncs}, last accessed 2023/10/25
% \end{thebibliography}

\end{document}